\documentclass{article}

    \usepackage[preprint]{neurips_2025}

\usepackage[utf8]{inputenc} 
\usepackage[T1]{fontenc}    
\usepackage{hyperref}       
\usepackage{url}            
\usepackage{booktabs}       
\usepackage{amsfonts}       
\usepackage{nicefrac}       
\usepackage{microtype}      
\usepackage{xcolor}         
\usepackage{array}
\usepackage[linesnumbered,ruled,vlined]{algorithm2e}
\usepackage{amsmath}
\usepackage{amsmath}
\usepackage{makecell}
\usepackage{multirow}
\usepackage{pifont}
\usepackage{bm}
\usepackage{graphicx}
\usepackage{colortbl}
\title{SRKD: Towards Efficient 3D Point Cloud Segmentation via Structure- and Relation-aware Knowledge Distillation}

\author{
\begin{tabular}{c}
Yuqi Li\textsuperscript{1}\thanks{Equal Contribution.}, 
Junhao Dong\textsuperscript{2}\footnotemark[1],
Zeyu Dong\textsuperscript{1}, 
Chuanguang Yang\textsuperscript{1}\thanks{Corresponding Author: yangchuanguang@ict.ac,cn},
Zhulin An\textsuperscript{1},
Yongjun Xu\textsuperscript{1}
\end{tabular} \\
\textsuperscript{1}Institute of Computing Technology, Chinese Academy of Sciences \\
\textsuperscript{2}Nanyang Technological University.\\
}

\begin{document}

\maketitle

\begin{abstract}
3D point cloud segmentation faces practical challenges due to the computational complexity and deployment limitations of large-scale transformer-based models. To address this, we propose a novel \textbf{S}tructure- and \textbf{R}elation-aware \textbf{K}nowledge \textbf{D}istillation framework, named \textbf{SRKD}, that transfers rich geometric and semantic knowledge from a large frozen teacher model (>100M) to a lightweight student model (<15M). Specifically, we propose an affinity matrix-based relation alignment module, which distills structural dependencies from the teacher to the student through point-wise similarity matching, enhancing the student’s capability to learn contextual interactions. Meanwhile, we introduce a cross-sample mini-batch construction strategy that enables the student to perceive stable and generalized geometric structure. This aligns across diverse point cloud instances of the teacher, rather than within a single sample.  Additionally, KL divergence is applied to align semantic distributions, and ground-truth supervision further reinforces accurate segmentation. Our method achieves state of the art performance with significantly reduced model complexity, demonstrating its effectiveness and efficiency in real-world deployment scenarios. Our Code is available at \url{https://github.com/itsnotacie/SRKD}.
\end{abstract}

\section{Introduction}
3D point cloud semantic segmentation has become a fundamental task in autonomous driving \cite{sun2025view}, robotics \cite{nguyen20133d}, and AR/VR \cite{kharroubi2019classification}, as it enables fine-grained scene understanding by assigning semantic labels to irregular and sparse 3D points. Compared to 2D images, point clouds naturally encode rich geometric and spatial information, making them highly suitable for perceiving object shapes, positions, and boundaries in real-world environments \cite{wu2019pointconv}. However, due to the inherent sparsity, unstructured nature, and permutation invariance of point clouds, achieving high segmentation accuracy while maintaining computational efficiency remains a significant challenge \cite{hou2022point}.

Recently, Transformer-based models have shown promising results in point cloud segmentation due to their powerful global context modeling capabilities \cite{zhao2021point, choy20194d, lai2022stratified, wang2023octformer, wu2022point, wu2024point, qu2024end}. Nevertheless, such models typically involve hundreds of millions of parameters and quadratic computational complexity, which pose serious challenges for real-time inference and edge deployment. Knowledge distillation (KD), formulated by Hinton \textit{et al.}~\cite{hinton2015distilling}, is an effective way for model compression. The core idea is to transfer meaningful knowledge from a cumbersome teacher to a lightweight student, therefore improving the student performance while retaining the lightweight property. KD has been applied to 2D vision tasks through distilling class probability~\cite{hinton2015distilling,yang2021hierarchical}, feature maps~\cite{romero2014fitnets,chen2021distilling,yang2024clip}, and relationships~\cite{park2019relational,yang2022mutual,yang2022cross,yang2023online,feng2024relational}. Although KD has proven to be an effective model compression technique in 2D vision tasks \cite{liu2019structured, yang2022cross, liu2024transkd}, its direct application to 3D point clouds is non-trivial. Compared to the traditional 2D feature maps, the disorderly and unstructured nature of 3D point clouds make it difficult to directly transfer knowledge from teacher to student without structural misalignment or loss of spatial fidelity though 2D related KD strategies \cite{hou2022point, zhang2023pointdistiller, su2025balanced}.

Furthermore, existing point cloud KD methods primarily focus on within-sample distillation, where the student learns from the teacher's predictions or features on a per-sample basis. However, this overlooks the potential cross-sample structural consistency and geometric interactions that exist across different point cloud instances. In practice, point clouds often exhibit recurring geometric patterns and shared contextual semantics, especially in urban scenes. Failing to exploit such inter-sample dependencies limits the student's ability to generalize structural representations and capture contextual relationships between points effectively.

To address these limitations, we propose a novel \textbf{S}tructure- and \textbf{R}elation-aware \textbf{K}nowledge \textbf{D}istillation framework, termed SRKD, which enhances student model performance through cross-sample geometric alignment and relation-aware interaction modeling. Specifically, we introduce an affinity matrix-based relation alignment module, which transfers point-wise interaction knowledge by aligning pairwise similarities between teacher and student features. Under the joint distillation strategy of multi-sample geometric and single-sample semantic guidance, SRKD enables the student model to effectively capture the core properties of the teacher model. Furthermore, we design a cross-sample mini-batch construction strategy that allows the student to absorb generalized geometric structures distilled from the frozen teacher model (>100M parameters). In addition, KL divergence is employed to align semantic distributions, and supervised learning with ground-truth labels further strengthens semantic consistency. Under the collaborative supervision of both soft labels and hard labels, SRKD achieves competitive performance with significantly reduced model complexity, demonstrating its practicality for real-world 3D segmentation applications.

Our key contributions can be summarized as: (1) We propose a novel Structure- and Relation-aware Knowledge Distillation (SRKD) framework, which significantly improves the performance of the student model in 3D segmentation tasks by incorporating cross-sample geometric alignment and relation-aware modeling. (2) We introduce an affinity matrix-based relation alignment module to transfer point-wise interaction knowledge at the feature level. In addition, we design a cross-sample mini-batch construction strategy that enables the student model to learn generalized geometric structures embedded in the high-capacity teacher model. (3) Extensive experiments on large-scale indoor and outdoor benchmarks validate the effectiveness of SRKD, demonstrating its superior performance and strong robustness across diverse 3D segmentation scenarios.

\section{Related Works}

\textbf{Learnable Point Cloud Semantic Segmentation.} Benefiting from the powerful data-driven capabilities of deep learning, it has become feasible to directly extract effective features from raw point clouds \cite{qi2017pointnet, qi2017pointnet++, choy20194d, thomas2019kpconv}. Inspired by this progress, a wide range of point cloud semantic segmentation methods have emerged. Early approaches, such as PointNet \cite{qi2017pointnet} and its variants \cite{qi2017pointnet++}, introduced the use of MLPs to process point clouds directly, enabling effective learning on small-scale datasets. However, these models exhibit limited scalability when applied to large-scale outdoor LiDAR scenes. To address this challenge, some researchers focus on large-scale scene segmentation \cite{landrieu2018large, hu2020randla, fan2021scf}. For example, \cite{hu2020randla} utilized random sampling and local feature aggregation modules to preserve critical geometric information, while \cite{xu2021rpvnet} proposed a range-point-voxel fusion strategy to leverage multiple representations. \cite{zhu2021cylindrical} exploited cylindrical partitioning and asymmetric convolutions to better capture the spatial structure. Nonetheless, the limited receptive field of many of these methods constrains their ability to model long-range dependencies, thereby impeding further improvements in segmentation accuracy. Transformers \cite{vaswani2017attention}, originally developed for natural language processing, possess an inherent capacity for capturing global context and have recently achieved remarkable success in point cloud segmentation tasks \cite{guo2021pct, zhao2021point, wu2022point, wu2024point}. 

Although these methods exhibit the strong performance on standard benchmarks, they typically incur substantial computational costs, such as Transformers exhibit quadratic complexity with respect to input size, posing significant challenges for deployment on resource-constrained devices. To bridge the gap between performance and efficiency, in this paper, we proposes a geometry- and relation-aware knowledge distillation framework, offering a feasible solution for compressing complex LiDAR segmentation models while maintaining high accuracy.

\textbf{Knowledge Distillation for 3D Tasks:} Knowledge Distillation (KD) aims to transfer rich and implicit knowledge from a large and cumbersome teacher model to a lightweight and compact student model, with the goal of narrowing the performance gap between them \cite{hinton2015distilling}. A large of works has demonstrated its practical value in real-world deployment scenarios. KD supports various forms of knowledge as distillation targets and has achieved remarkable success in image classification tasks \cite{hou2019learning, romero2014fitnets, hou2019learning, park2019relational, yim2017gift}. Motivated by the pursuit of more efficient segmentation, researchers have extended KD to the field of image segmentation. Inspired by its effectiveness in 2D vision tasks, KD has also been widely adopted in a range of 3D applications. Several studies have explored the application of KD to point cloud object detection \cite{yang2022towards, klingner2023x3kd, zhang2023pointdistiller}, significantly improving detection efficiency. More recently, several works have attempted to employ KD in point cloud semantic segmentation \cite{hou2022point, yang2023label}, showcasing its promising potential in 3D scene understanding tasks.

Although several knowledge distillation (KD) approaches have achieved promising results in 3D semantic segmentation tasks, they often suffer from either a lack of cross-sample geometric awareness or insufficient alignment of contextual relationships. In this paper, we propose a knowledge distillation framework that effectively transfers semantic knowledge from a powerful teacher model to a lightweight student model. Specifically, we introduce a cross-sample construction strategy, enabling the student model to capture the geometric structures across multiple samples as represented by the teacher. Additionally, a context relation matrix is designed to guide point-wise semantic alignment between the student and teacher models.

\section{Methodology}

In this section, we first introduce the problem formulation of point cloud semantic segmentation. Then, a cross-sample mini-batch strategy and an affinity matrix-based relation alignment module are presented. Finally, we describe the overall distillation framework along with the training and inference procedures.

\begin{algorithm}[htbp]
\scriptsize
\DontPrintSemicolon
\SetAlgoLined
\KwIn{Training set $\mathcal{D}$; Frozen teacher model $f_t$; Student model $f_s$ with parameters $\theta_s$}
\KwOut{Optimized student model parameters $\theta_s$}

\For{each mini-batch $B = \{(P_b, y_b)\}_{b=1}^{B}$}{
    $Z_t, F_t \leftarrow f_t(B)$

    $Z_s, F_s \leftarrow f_s(B)$

    $M_{ij}^t \leftarrow \text{ComputeCrossSampleSimilarity}(F_t)$

    $M_{ij}^s \leftarrow \text{ComputeCrossSampleSimilarity}(F_s)$

    $L_{\text{batch-GD}} \leftarrow \frac{1}{N^2} \sum_{i,j} KL\left(\sigma(M_{ij}^s) \parallel \sigma(M_{ij}^t)\right)$

    $D_p^t \leftarrow \text{ComputePointAffinity}(F_t)$

    $D_p^s \leftarrow \text{ComputePointAffinity}(F_s)$

    $L_{\text{amra}_p} \leftarrow \frac{1}{N_p^2} \sum_{i,j} \left\| D_p^s(i,j) - D_p^t(i,j) \right\|^2$

    $D_v^t \leftarrow \text{ComputeVoxelAffinity}(F_t)$

    $D_v^s \leftarrow \text{ComputeVoxelAffinity}(F_s)$

    $L_{\text{amra}_v} \leftarrow \frac{1}{N_v^2} \sum_{i,j} \left\| D_v^s(i,j) - D_v^t(i,j) \right\|^2$

    $L_{\text{amra}_c} \leftarrow \frac{1}{N_p} \sum_i KL\left(\sigma(F_s^{(i)}) \parallel \sigma(F_t^{(i)})\right)$

    $L_{\text{kd}} \leftarrow \frac{1}{N} \sum_i KL\left(\sigma(Z_s^{(i)}) \parallel \sigma(Z_t^{(i)})\right)$

    $L_{\text{task}} \leftarrow \frac{1}{N} \sum_i CE\left(\sigma(Z_s^{(i)}), y^{(i)}\right)$

    $L_{\text{total}} \leftarrow L_{\text{task}} + L_{\text{kd}} + L_{\text{amra}_p} + L_{\text{amra}_v} + L_{\text{amra}_c} + L_{\text{batch-GD}}$

    $\theta_s \leftarrow \theta_s - \eta \nabla_{\theta_s} L_{\text{total}}$
}

$\hat{y} \leftarrow f_s(P_{\text{test}})$

\caption{SRKD: Structure- and Relation-aware Knowledge Distillation}
\end{algorithm}

\subsection{Problem Setting}
Unlike conventional point cloud classification task, point cloud semantic segmentation involves a finer-grained  process. Generally, the segmentation network is required to predict the semantic category for each point in the point cloud from a predefined set of $C$ classes. This per-point prediction demands that the output resolution be aligned with the input resolution, thus often resulting in a computationally heavy network architecture that presents challenges for real-world deployment.

A typical segmentation network consists of two main components: a feature extractor and a segmentation head. The feature extractor takes the raw point cloud $P \in \mathbb{R}^{N \times 3}$ as input and produces a dense feature map $F \in \mathbb{R}^{N \times D}$, while the segmentation head consumes this feature map and outputs the final semantic logits $Z \in \mathbb{R}^{N \times C}$ for each point. Conventional segmentation approaches typically employ cross-entropy loss as the objective function to optimize the network parameters:

\begin{equation}
\begin{split}
\label{f311}
    \mathcal{L}_{task}=\frac{1}{N}\sum_{i=1}^{N}CE(\sigma(Z_i),y_i),
\end{split}
\end{equation}

where $N$ denotes the number of points in the point cloud, $CE(\cdot)$ represents the cross-entropy function, $\sigma$ means the softmax function, and $y$ indicates the ground-truth semantic label.

Following Hinton’s knowledge distillation framework \cite{hinton2015distilling}, we employ KL divergence to align the per-point class probability distributions of the student and teacher models:

\begin{equation}
\begin{split}
\label{f312}
    \mathcal{L}_{kd}=\frac{1}{N}\sum_{i=1}^{N}KL(\frac{\sigma(Z_i^s)}{T}||\frac{\sigma(Z_i^t)}{T}),
\end{split}
\end{equation}

where $T$ is the temperature function (typically set to 1 for optimal performance \cite{liu2019structured}), and $\frac{\sigma(Z_i^s)}{T}$, $\frac{\sigma(Z_i^t)}{T}$ are the student and teacher models' soft class predictions respectively.

\subsection{Affinity Matrix-Based Relation Alignment}


Although the KL divergence computed on final predictions can enhance the effectiveness of knowledge distillation for the student model, it overlooks the semantic information embedded in intermediate feature representations. Meanwhile, extracting knowledge solely from point-wise outputs is insufficient, as it only captures individual semantic cues while failing to model the structural information of the surrounding context. For point cloud with the unordered nature, the contextual structural knowledge is particularly critical for LiDAR-based semantic segmentation models. To address this issue, we propose an efficient relational affinity distillation strategy that facilitates the learning of point-to-point relational knowledge in a scalable manner.

\textbf{Voxel Sampling.} We first  partition the entire point cloud into multiple voxels $R_v \times A_v \times H_v$. Then, each voxel is further divided into a number of sub-voxels. The total number of voxels is denoted as $N_v = \lceil \frac{R}{R_v} \rceil \times \lceil \frac{A}{A_v} \rceil \times \lceil \frac{H}{H_v} \rceil$. To facilitate relational affinity distillation, we adopt a class-aware sampling strategy that selects $K$ super-voxels. The sampling weight for the i-th super-voxel is defined as follows:

\begin{equation}
\begin{split}
\label{f331}
    w_i =  \frac{\tau_{class}}{N_v} \cdot \frac{D_i}{R},
\end{split}
\end{equation}

where $R$, $A$, and $H$ mean radius, angle, and hight, respectively. $\lceil \cdot \rceil$ denotes the ceiling function. $D_i$ is the distance from the outer contour of the i-th supervoxel to the origin of the XY plane. $\tau_{class}=1-\frac{C_{current}}{C_{total}}$, this achieves class balancing, where categories with fewer samples are assigned higher sampling weights, and vice versa

In practical applications, the input size and point density of each point cloud can vary significantly, resulting in a variable number of point features $N_{point}$ and non-empty voxels  $N_{voxel}$. To address this, we fix the number of point and voxel features by discarding excess elements when the count exceeds a predefined threshold, and padding with zeros when it falls below the threshold.

\textbf{Calculate affinity matrix.} Let $\mathbf{F}_{point} \in \mathbb{R}^{N_p \times D_p}$ denote the $N_{point}$ point features and $\mathbf{F}_v \in \mathbb{R}^{N_v \times D_v}$ denote the $N_{voxel}$ voxel features within the r-th supervoxel. For each supervoxel, we compute the inter-point affinity matrix based on the $\mathbb{L}_2$ distance:

\begin{equation}
\begin{split}
\label{f332}
    \mathcal{D}(i,j,w_i) =w_i||F^i-F^j||_2^2,
\end{split}
\end{equation}

The affinity scores capture the similarity between each pair of point features and serve as high-level structural knowledge for guiding the student model. The point-wise affinity distillation loss is defined as follows:

\begin{equation}
\begin{split}
\label{f333}
\mathcal{L}_{amra}^p= \frac{1}{N_{point}^2} \sum_{i=1}^{N_{point}} \sum_{j=1}^{N_{point}} ||\mathcal{D}_S^p(i,j,w_i) - \mathcal{D}_T^p(i,j,w_i)||_2^2.
\end{split}
\end{equation}

The computation of the inter-voxel affinity matrix follows a similar procedure. Ultimately, the student model is guided to mimic the affinity matrices generated by the teacher model. The inter-voxel affinity distillation loss is formulated as follows:

\begin{equation}
\begin{split}
\label{f334}
\mathcal{L}_{amra}^v= \frac{1}{N_{voxel}^2}  \sum_{i=1}^{N_{voxel}} \sum_{j=1}^{N_{voxel}} \|\mathcal{D}_S^v(i,j,w_i) - \mathcal{D}_T^v(i,j,w_i)\|_2^2.
\end{split}
\end{equation}

Furthermore, to better exploit the knowledge encoded in each channel, we propose a soft alignment of the activations between corresponding channels in the teacher and student networks. The channel-wise distillation loss can be formulated in a general form as follows:

\begin{equation}
\begin{split}
\label{f335}
\mathcal{L}_{amra}^c=\quad\quad\quad\quad\quad\quad\quad\quad\quad\quad\quad\quad\quad\quad\quad\quad\quad\quad\quad\quad\quad\quad\quad\quad\quad\quad\quad\quad\quad\quad\quad\quad\quad\quad\quad\\
\frac{1}{N_{point}} \sum_{i=1}^{N_{point}}KL(\sigma(F_{point}^S)||\frac{1}{N_{point}} (\sigma(F_{point}^T))+\sum_{i=1}^{N_{voxel}}KL(\sigma(F_{voxel}^S)||(\sigma(F_{voxel}^T))
\end{split}
\end{equation}

\subsection{Cross-Sample Mini-Batch Geometry Distillation}
Existing 3D semantic segmentation methods typically guide the student model to learn geometric representations generated by the teacher model from a single point cloud sample. However, unlike images with regular grid structures, point clouds are inherently unordered and unstructured (each point cloud has a different number and density of points), making it difficult for such methods to capture geometric similarities and variations across different samples. However, these global geometric semantics provide critical representational knowledge for understanding point clouds. In this paper, we propose a cross-sample mini-batch geometric alignment strategy that enables the student model to learn the general geometric semantics encoded by the teacher model across multiple point clouds.

\textbf{Geometric similarity matrix.} Given a mini-batch $\{P_b\}_{b=1}^{B}$ consisting of $B$ point clouds, the segmentation network encodes each input into a corresponding geometric feature map $F_b \in \mathbb{R}^{N \times D}$. Then, $\mathcal{L}_2$ normalization is applied to the point-wise feature embeddings within each feature map $F_b$ to ensure consistent scale and comparability. To capture geometric knowledge across different samples, we compute a cross-sample similarity matrix $M_{ij}=F_iF_j^T, 1 \leq i \leq B, 1 \leq j \leq B$, between the i-th and j-th point clouds in the mini-batch, which reflects the pairwise geometric similarity between their respective points.

\textbf{Geometric knowledge alignment.} To facilitate knowledge transfer, we enforce the student’s pairwise similarity matrix $M_{ij}^s$ to be consistent with the counterpart  $M_{ij}^t$ computed by the teacher. Therefore, the distillation objective is defined as follows:

\begin{equation}
\begin{split}
\label{f321}
    \mathcal{L}_{\text{GD}}(\mathbf{M}^{M}_{ij}, \mathbf{S}^{t}_{ij}) = \frac{1}{N} \sum_{a=1}^{N} \mathrm{KL} \left( \sigma \left( \frac{\mathbf{M}^{s}_{ij}[a,:]}{T} \right) \Big\| \sigma \left( \frac{\mathbf{M}^{t}_{ij}[a,:]}{T} \right) \right),
\end{split}
\end{equation}
where $\mathbf{M}^{s}_{ij}[a,:]$ means the a-th row vector of $\mathbf{M}^{s}_{ij}$.

We normalize each row of the similarity matrix $M_{ij}$ into a temperature-scaled probability distribution using the softmax function $\sigma$ with temperature parameter $T$. The softmax normalization eliminates magnitude discrepancies between the student and teacher networks. The KL divergence is then employed to align the probability distributions row-wise. Our approach performs point-to-point distillation for every two point clouds in the mini-batch:

\begin{equation}
\begin{split}
\label{f322}
    \mathcal{L}_{\text{batch-GD}} = \frac{1}{N^2} \sum_{i=1}^{N} \sum_{j=1}^{N} \mathcal{L}_{\text{GD}}(\mathbf{M}^{s}_{ij}, \mathbf{M}^{t}_{ij}).
\end{split}
\end{equation}

\subsection{Student-Teacher Distillation}

\begin{figure*}[htp]
	\centering
	\includegraphics[width=\textwidth]
 {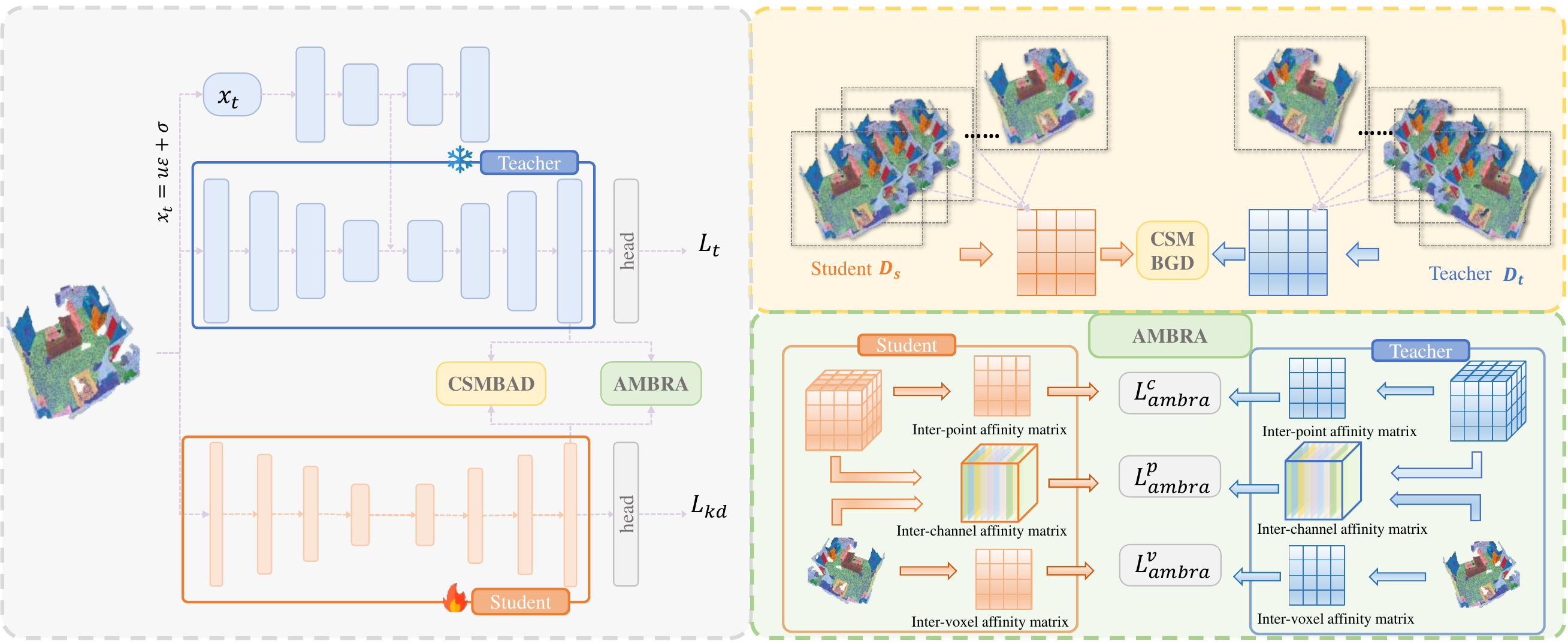}
 \vspace{-0.4cm} 
	\caption{Overall Frameworkof SRKD. SRKD enables the student model to capture both coarse-grained and fine-grained semantic knowledge from the teacher model by leveraging geometric structural cues and semantic relational information at multiple representation levels.}
	\label{fig1}
\end{figure*}

Our overall framework is simply and clearly described in Fig.~\ref{fig1}.

\textbf{Teacher Model.} For the teacher network, we adopt CDSegNet \cite{qu2024end} due to its outstanding performance on point cloud semantic segmentation tasks. CDSegNet is a dual-branch architecture based on denoising diffusion probabilistic models (DDPMs) \cite{ho2020denoising}. The conditional branch, built upon PTv3 as the backbone, dominates the final segmentation output. Meanwhile, a lightweight denoising network is incorporated as an auxiliary branch to enhance robustness. Although the denoising branch is compact, the overall CDSegNet still contains over 100 million parameters, which poses significant challenges for real-world deployment.

\textbf{Student Model.} To align with the teacher model, we employ a student model derived from PTv3 \cite{wu2024point} with channel dimensions reduced by half, resulting in a parameter count that is less than 1/8 of CDSegNet. The student architecture follows the structure of the conditional branch in CDSegNet, which is responsible for generating the primary segmentation output.

\textbf{Distillation Process.} The input point clouds are simultaneously passed through the frozen teacher model and the student model to extract features $F_t$ and $F_s$, respectively. First, the student model is guided to mimic the cross-sample geometric knowledge encoded by the teacher using Equation (3). Then, to further capture unordered structural relationships, voxel-to-point relation-aware distillation is applied via Equation (5). Finally, a hybrid supervision scheme is adopted, combining ground-truth labels with the soft labels generated by the teacher, further enhancing the segmentation performance of the student model.

\subsection{Training and Inference.}

\textbf{Training.} As described in the previous sections, we impose multiple constraints on the student model through semantic segmentation distillation. Specifically, the training objective integrates cross-sample geometric alignment, relation-aware affinity matching, and hybrid supervision with both ground-truth and soft labels provided by the teacher model:

\begin{equation}
\begin{split}
\label{f322}
    \mathcal{L}_{total}=\mathcal{L}_{task}+\lambda_{kd}\mathcal{L}_{kd}+\lambda_{p}\mathcal{L}_{amra}^p+\lambda_{v}\mathcal{L}_{amra}^v+\lambda_{c}\mathcal{L}_{amra}^c+\lambda_{\text{batch-GD}}\mathcal{L}_{\text{batch-GD}}.
\end{split}
\end{equation}
Here, $\lambda_{kd}$, $\lambda_{p}$, $\lambda_{v}$, $\lambda_{c}$, and $\lambda_{\text{batch-GD}}$ are loss weights for balancing magnitudes of various losses.  

\textbf{Inference.} During inference, we follow the standard protocol of PTv3. Only the lightweight student model is used to perform semantic segmentation, ensuring efficient and real-time deployment without incurring additional computational overhead.

\section{Experiments}

\subsection{Experiment Setup}

\textbf{Dataset.} We evaluate our method on the widely used large-scale indoor dataset ScanNet \cite{dai2017scannet}. The dataset consists of 1,513 annotated 3D scenes. Following the official protocol, we use 1,201 scenes for training, 312 for validation, and 100 for testing. All point clouds in ScanNet are voxelized with a resolution of 0.02 meters.

\textbf{Baseline.} To validate the effectiveness of our distillation strategy, we adopt the PTv3 model with reduced channel width (without any distillation) as the baseline for comparison.

\textbf{Training Details.} All experiments are conducted on a single NVIDIA RTX 4090 GPU (24GB VRAM) with 12-core CPU and 29GB memory, running Ubuntu 20.04 and CUDA 11.8. The implementation is based on PyTorch 2.0.0, The models are trained for 800 epochs using the AdamW optimizer with an initial learning rate of 0.006, weight decay of 0.05, and a cosine learning rate schedule (OneCycleLR) with and . The batch size is set to 8, and mixed-precision training (AMP) is enabled to accelerate training. The KL divergence temperature is fixed to $T=2$. Loss components are weighted as follows: KL loss (semantic alignment) with weight $\lambda_{kd}=0.3$, point-level and inter-voxel affinity distillation loss with weight $\lambda_{p}=\lambda_{v}=0.001$, channel-wise distillation loss with weight $\lambda_{c}=1000$, and geometric similarity (MSE) loss with weight $\lambda_{\text{batch-GD}}=0.1$. The loss weights are empirically defined by balancing these losses to the same order of magnitudes.

\subsection{Comparison of segmentation results}

\begin{table}[h]
	\begin{center}
	\begin{tabular}{p{4cm}p{1cm}p{1cm}p{1cm}p{2.2cm}}	
        \Xhline{1pt}

        Method
        &\makecell[c]{mIoU}
        &\makecell[c]{mAcc}
        &\makecell[c]{allAcc}
        &\makecell[c]{$\#$Params}\\
       \hline

        MinkUNet \cite{choy20194d}
        &\makecell[c]{72.3}
        &\makecell[c]{79.4}
        &\makecell[c]{89.1}
        &\makecell[c]{37.9M}\\

        OctFormer \cite{wang2023octformer}
        &\makecell[c]{75.0}
        &\makecell[c]{83.1}
        &\makecell[c]{91.3}
        &\makecell[c]{44.0M}\\

        PTv2 \cite{wu2022point}
        &\makecell[c]{75.5}
        &\makecell[c]{82.9}
        &\makecell[c]{91.2}
        &\makecell[c]{12.8M}\\
        
        PTv3 \cite{wu2024point}
        &\makecell[c]{\underline{77.6}}
        &\makecell[c]{85.0}
        &\makecell[c]{{92.0}}
        &\makecell[c]{46.2M}\\

        CDSegNet \cite{qu2024end}
        &\makecell[c]{\textbf{77.9}}
        &\makecell[c]{\underline{85.2}}
        &\makecell[c]{{92.2}}
        &\makecell[c]{101.4M}\\

        Baseline
        &\makecell[c]{76.7}
        &\makecell[c]{84.2}
        &\makecell[c]{91.6}
        &\makecell[c]{11.6M}\\

        \rowcolor{gray!20} 
        Ours
        &\makecell[c]{\textbf{77.9}}
        &\makecell[c]{\textbf{85.7}}
        &\makecell[c]{\underline{92.1}}
        &\makecell[c]{11.6M}\\

        \Xhline{1pt}
        
	\end{tabular}
	\end{center}
 
	\caption{The results on ScanNet. Our method achieves state-of-the-art performance with minimal number of parameters.}
	\label{tab421}

\end{table}

\begin{figure*}[htp]
	\centering
	\includegraphics[width=\textwidth]
 {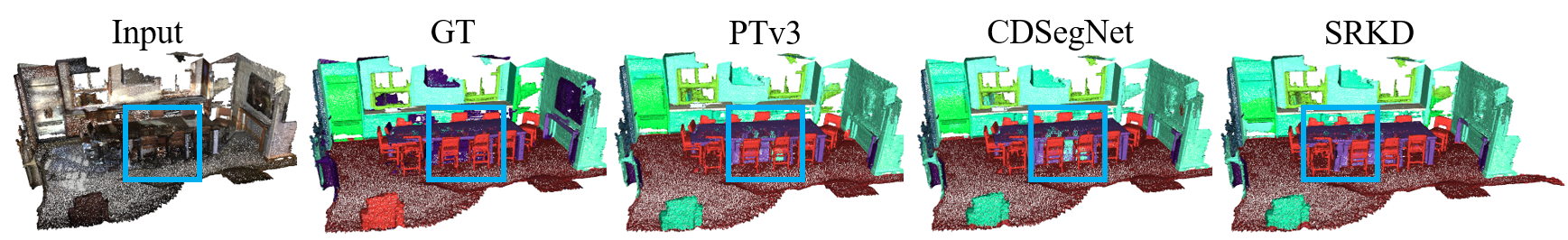}
 \vspace{-0.4cm} 
	\caption{The visualization results on ScanNet.}
	\label{fig1}
\end{figure*}

\textbf{Indoor Dataset.} We first compare our method with existing state-of-the-art point cloud semantic segmentation methods. As shown in Tab.~\ref{tab421}, our method achieves superior segmentation performance, demonstrating its effectiveness. These results validate that SRKD enables the student model to effectively acquire semantic knowledge from the teacher model through the joint optimization of geometry- and relation-aware soft labels together with ground-truth supervision even when the student model has less than 1/8 the number of parameters of the teacher model.

\begin{table*}[h]
  \resizebox{1.0\textwidth}{!}{
        \begin{tabular}{p{2.9cm}|p{0.9cm}|p{1.0cm}p{1.1cm}p{0.9cm}p{0.9cm}p{1.2cm}p{0.9cm}p{1.5cm}p{0.9cm}p{1.0cm}p{1.0cm}p{1.2cm}p{1.0cm}p{1.3cm}p{1.0cm}p{1.4cm}p{1.5cm}}	
        \Xhline{1pt}
  
        {Methods}
        &\makecell[c]{mIoU}
        &{\makecell[c]{barrier}}
        &{\makecell[c]{bicycle}}
        &{\makecell[c]{bus}}
        &{\makecell[c]{car}}
        &{\makecell[c]{vehicle}}
        &{\makecell[c]{moto}}
        &{\makecell[c]{pedestrian}}
        &{\makecell[c]{cone}}
        &{\makecell[c]{trailer}}
        &{\makecell[c]{truck}}
        &{\makecell[c]{drivable}}
        &{\makecell[c]{others}}
        &{\makecell[c]{sidewalk}}
        &{\makecell[c]{terrain}}
        &{\makecell[c]{manmade}}
        &{\makecell[c]{vegetation}} \\
        
       \hline

        Cylinder3D \cite{zhu2021cylindrical}
        &\makecell[c]{76.1}
        &\makecell[c]{76.4}
        &\makecell[c]{40.3}
        &\makecell[c]{91.2}
        &\makecell[c]{{93.8}}
        &\makecell[c]{51.3}
        &\makecell[c]{78.0}
        &\makecell[c]{78.9}
        &\makecell[c]{64.9}
        &\makecell[c]{62.1}
        &\makecell[c]{84.4}
        &\makecell[c]{96.8}
        &\makecell[c]{71.6}
        &\makecell[c]{76.4}
        &\makecell[c]{75.4}
        &\makecell[c]{90.5}
        &\makecell[c]{87.4}
        \\

        PVKD \cite{hou2022point}
        &\makecell[c]{76.0}
        &\makecell[c]{76.2}
        &\makecell[c]{40.0}
        &\makecell[c]{90.2}
        &\makecell[c]{\underline{94.0}}
        &\makecell[c]{50.9}
        &\makecell[c]{77.4}
        &\makecell[c]{78.8}
        &\makecell[c]{64.7}
        &\makecell[c]{62.0}
        &\makecell[c]{84.1}
        &\makecell[c]{96.6}
        &\makecell[c]{71.4}
        &\makecell[c]{76.4}
        &\makecell[c]{\underline{76.3}}
        &\makecell[c]{90.3}
        &\makecell[c]{86.9}
        \\

        PTv3 \cite{wu2024point}
        &\makecell[c]{80.3}
        &\makecell[c]{\textbf{80.5}}
        &\makecell[c]{\textbf{53.8}}     
        &\makecell[c]{\underline{95.9}}
        &\makecell[c]{91.9}
        &\makecell[c]{52.1}
        &\makecell[c]{\underline{88.9}}
        &\makecell[c]{\textbf{84.5}}
        &\makecell[c]{\textbf{71.7}}
        &\makecell[c]{\textbf{74.1}}
        &\makecell[c]{{84.5}}
        &\makecell[c]{\underline{97.2}}
        &\makecell[c]{\underline{75.6}}
        &\makecell[c]{77.0}
        &\makecell[c]{{76.2}}
        &\makecell[c]{\underline{91.2}}
        &\makecell[c]{\underline{89.6}}  \\

        CDSegNet \cite{qu2024end}
        &\makecell[c]{\textbf{81.2}}
        &\makecell[c]{\underline{80.1}}
        &\makecell[c]{\underline{53.5}}        
        &\makecell[c]{\textbf{97.0}}
        &\makecell[c]{92.3}
        &\makecell[c]{\textbf{62.3}}
        &\makecell[c]{\textbf{89.7}}
        &\makecell[c]{\underline{84.2}}
        &\makecell[c]{\textbf{71.7}}
        &\makecell[c]{72.2}
        &\makecell[c]{\underline{85.9}}
        &\makecell[c]{\underline{97.2}}
        &\makecell[c]{\textbf{76.5}}
        &\makecell[c]{\underline{77.8}}
        &\makecell[c]{\textbf{76.9}}
        &\makecell[c]{\textbf{91.4}}
        &\makecell[c]{\textbf{89.7}}  \\

        Baseline
        &\makecell[c]{77.9}
        &\makecell[c]{77.4}
        &\makecell[c]{49.3}        
        &\makecell[c]{93.7}
        &\makecell[c]{91.6}
        &\makecell[c]{51.2}
        &\makecell[c]{87.0}
        &\makecell[c]{80.8}
        &\makecell[c]{68.2}
        &\makecell[c]{70.0}
        &\makecell[c]{83.7}
        &\makecell[c]{95.0}
        &\makecell[c]{74.2}
        &\makecell[c]{74.4}
        &\makecell[c]{73.9}
        &\makecell[c]{88.6}
        &\makecell[c]{87.4}  \\

        \rowcolor{gray!20} 
        Ours
        &\makecell[c]{80.5}
        &\makecell[c]{79.6}
        &\makecell[c]{51.5}        
        &\makecell[c]{95.1}
        &\makecell[c]{\textbf{94.4}}
        &\makecell[c]{\underline{61.6}}
        &\makecell[c]{85.6}
        &\makecell[c]{82.7}
        &\makecell[c]{\underline{70.7}}
        &\makecell[c]{\underline{73.5}}
        &\makecell[c]{\textbf{86.9}}
        &\makecell[c]{97.0}
        &\makecell[c]{76.2}
        &\makecell[c]{\underline{77.1}}
        &\makecell[c]{76.2}
        &\makecell[c]{91.0}
        &\makecell[c]{89.4}  \\
        
        \Xhline{1pt}
        
	\end{tabular}
 }
	\caption{The results on nuScenes. }
	\label{tab522}
\end{table*}

\textbf{Outdoor Dataset.} We further validate our approach on an outdoor benchmark. Compared to indoor environments, outdoor scenes typically exhibit greater spatial distances between points, resulting in increased data sparsity. This characteristic poses additional challenges for accurate geometric understanding and feature association. Table 3 presents the superior performance of SRKD, which significantly outperforms the baseline. By enabling feature-level knowledge alignment across point-to-voxel and channel dimensions, as well as multi-sample geometric structure alignment, SRKD allows the student model to deeply capture the intrinsic properties of the teacher model. This leads to remarkable results on the challenging task of segmentation in sparse radar scenarios.

\subsection{Noise Experiments}

\begin{table}[h]
        
	\begin{center}
	\begin{tabular}{p{3.5cm}p{1.1cm}p{1.1cm}p{1.2cm}p{0.1cm}p{1.1cm}p{1.1cm}p{1.2cm}}	
        \Xhline{1pt}
  
        \multirow{2}{*}{Methods}
        &\multicolumn{3}{c}{Smalle $\tau$ (mIoU)} 
        &\quad
        &\multicolumn{3}{c}{Big $\tau$ (mIoU) } \\
         \cline{2-4} \cline{6-8}
        
        &\makecell[c]{$\bm{\tau}$=0.01}
        &\makecell[c]{$\bm{\tau}$=0.05}
        &\makecell[c]{$\bm{\tau}$=0.1}
        &\quad
        &\makecell[c]{$\bm{\tau}$=0.5}
        &\makecell[c]{$\bm{\tau}$=0.7}
        &\makecell[c]{$\bm{\tau}$=1.0}\\
        
        \hline

        PTv2 \cite{wu2022point}
        &\makecell[c]{75.5}
        &\makecell[c]{75.4}
        &\makecell[c]{73.7}
        &\quad
        &\makecell[c]{8.7}
        &\makecell[c]{1.5}
        &\makecell[c]{1.2}\\

        PTv3 \cite{wu2024point}
        &\makecell[c]{\underline{77.6}}
        &\makecell[c]{\underline{77.4}}
        &\makecell[c]{\underline{76.9}}
        &\quad
        &\makecell[c]{{45.8}}
        &\makecell[c]{{26.0}}
        &\makecell[c]{{12.9}}\\

        CDSegNet \cite{qu2024end}
        &\makecell[c]{\textbf{77.9}}
        &\makecell[c]{\textbf{77.7}}
        &\makecell[c]{\textbf{77.2}}
        &\quad
        &\makecell[c]{\textbf{57.0}}
        &\makecell[c]{\textbf{46.7}}
        &\makecell[c]{\textbf{35.9}}\\

        Baseline
        &\makecell[c]{76.5}
        &\makecell[c]{76.3}
        &\makecell[c]{76.0}
        &\quad
        &\makecell[c]{56.8}
        &\makecell[c]{46.1}
        &\makecell[c]{25.6}\\

        \rowcolor{gray!20} 
        Ours
        &\makecell[c]{76.6}
        &\makecell[c]{76.4}
        &\makecell[c]{76.2}
        &\quad
        &\makecell[c]{\textbf{57.1}}
        &\makecell[c]{\underline{46.2}}
        &\makecell[c]{\underline{25.9}}\\
        
        \Xhline{1pt}
        
	\end{tabular}
	\end{center}
	\caption{The results of multiple distribution noise robustness on ScanNet. Our method can effectively exploit the noise robustness of the teacher model (CDSegNet).}
	\label{tab431}

\end{table}

Since CDSegNet leverages DDPMs to enhance robustness against noise, we further conduct noise robustness experiments to validate the effectiveness of SRKD. Specifically, we introduce Gaussian perturbations to the regularized feature representations of the model, i.e. $\bm{n_G} \sim \mathcal{N}(\bm{n_G};\bm{0},\tau\bm{I})$. As shown in Tab.~\ref{tab431}, SRKD demonstrates strong robustness under noisy conditions, even surpassing CDSegNet when the variance is set to $\tau=0.5$. These results indicate that SRKD can effectively exploit geometric and relation-aware knowledge distilled from the teacher model, enabling the student model to inherit key properties of the teacher.

\subsection{Subsampled Data}

\begin{table}[h]
	\begin{center}
	\begin{tabular}{p{2.8cm}p{1cm}p{1cm}p{1cm}p{1cm}p{1cm}p{1cm}}
        \Xhline{1pt}

        Method
        &\makecell[c]{$100\%$}
        &\makecell[c]{$50\%$}
        &\makecell[c]{$25\%$}
        &\makecell[c]{$12.5\%$}
        &\makecell[c]{$10\%$}
        &\makecell[c]{$5\%$}
        \\
       \hline

        PTv1 \cite{zhao2021point}
        &\makecell[c]{70.8}
        &\makecell[c]{63.4}
        &\makecell[c]{50.1}
        &\makecell[c]{32.7}
        &\makecell[c]{21.1}
        &\makecell[c]{12.9}
        \\

        MinkUNet \cite{choy20194d}
        &\makecell[c]{72.3}
        &\makecell[c]{67.5}
        &\makecell[c]{55.9}
        &\makecell[c]{45.3}
        &\makecell[c]{40.9}
        &\makecell[c]{28.5}\\

        ST \cite{lai2022stratified}
        &\makecell[c]{74.3}
        &\makecell[c]{69.0}
        &\makecell[c]{56.2}
        &\makecell[c]{49.7}
        &\makecell[c]{43.7}
        &\makecell[c]{30.8}\\

        OctFormer \cite{wang2023octformer}
        &\makecell[c]{75.0}
        &\makecell[c]{69.9}
        &\makecell[c]{56.5}
         &\makecell[c]{51.0}
        &\makecell[c]{45.9}
        &\makecell[c]{31.0}\\
        
        PTv2 \cite{wu2022point}
        &\makecell[c]{75.5}
        &\makecell[c]{70.3}
        &\makecell[c]{55.1}
        &\makecell[c]{50.6}
        &\makecell[c]{42.2}
        &\makecell[c]{29.5}\\

        PTv3 \cite{wu2024point}
        &\makecell[c]{\underline{77.6}}
        &\makecell[c]{73.0}
        &\makecell[c]{64.9}
        &\makecell[c]{56.7}
        &\makecell[c]{50.4}
        &\makecell[c]{41.5}\\

        CDSegNet \cite{qu2024end}
        &\makecell[c]{\textbf{77.9}}
        &\makecell[c]{\underline{73.9}}
        &\makecell[c]{\underline{66.5}}
        &\makecell[c]{\underline{64.5}}
        &\makecell[c]{\underline{62.3}}
        &\makecell[c]{\underline{46.2}}\\

        Baseline
        &\makecell[c]{76.7}
        &\makecell[c]{73.2}
        &\makecell[c]{61.2}
         &\makecell[c]{64.1}
        &\makecell[c]{62.2}
        &\makecell[c]{34.0}\\

        \rowcolor{gray!20} 
        Ours
        &\makecell[c]{\textbf{77.9}}
        &\makecell[c]{\textbf{74.0}}
        &\makecell[c]{\textbf{68.4}}
        &\makecell[c]{\textbf{65.4}}
        &\makecell[c]{\textbf{62.4}}
        &\makecell[c]{\textbf{57.1}}\\

        \Xhline{1pt}
        
	\end{tabular}
	\end{center}
 
	\caption{The results on ScanNet. Our method surpasses other methods in almost all metrics.}
	\label{tab521}

\end{table}

In addition, we evaluate the robustness of our method under data subsampling conditions. Specifically, we randomly sample $5\%$, $10\%$, $12.5\%$, $25\%$, and $50\%$ of the training and validation scenes from the ScanNet dataset. We then train and fine-tune our model on these subsampled datasets and evaluate its performance on the full validation set. Similar to the noise robustness experiments, SRKD is able to distill the teacher model’s robustness to sparse data, achieving remarkable performance under sub-sampled training data.

\subsection{Ablation Study}

To validate the effectiveness of our components, we primarily evaluate the cross-sample mini-batch construction strategy and the affinity matrix-based relation alignment module.

\begin{table}[h]
	\begin{center}
        \begin{tabular}{m{5.8cm}m{0.8cm}m{0.8cm}m{0.8cm}m{1.2cm}}
        \Xhline{1pt}
        Method
        &\makecell[c]{mIoU}
        &\makecell[c]{mAcc}
        &\makecell[c]{allAcc}\\
       \hline

        Baseline
        &\makecell[c]{70.8}
        &\makecell[c]{76.4}
        &\makecell[c]{87.5}\\

        Baseline+$\mathcal{L}_kd$
        &\makecell[c]{72.3}
        &\makecell[c]{79.4}
        &\makecell[c]{89.1}\\

        Baseline+$\mathcal{L}_kd$+CSMBGD
        &\makecell[c]{74.3}
        &\makecell[c]{82.5}
        &\makecell[c]{90.7}\\

        \rowcolor{gray!20} Baseline+$\mathcal{L}_kd$+CSMBGD+AMBRA 
        &\makecell[c]{\textbf{75.0}}
        &\makecell[c]{\textbf{83.1}}
        &\makecell[c]{\textbf{91.3}}\\

        \Xhline{1pt}
        
	\end{tabular}
	\end{center}
 
	\caption{The ablation study of module effectiveness.}
	\label{tab451}
\vspace{-0.5cm}
\end{table}

\textbf{} We first conduct an overall ablation study to evaluate the effectiveness of our proposed modules. The results are presented in Tab.~\ref{tab451}. Here, Baseline refers to the channel-reduced PTv3 model without any distillation. As shown, the performance of the baseline drops significantly without distillation. Furthermore, progressively removing the cross-sample mini-batch construction strategy and the affinity matrix-based relation alignment module leads to a gradual decline in segmentation performance, indicating the critical role of each component.

\begin{figure*}[htp]
	\centering
	\includegraphics[width=0.9\textwidth]
 {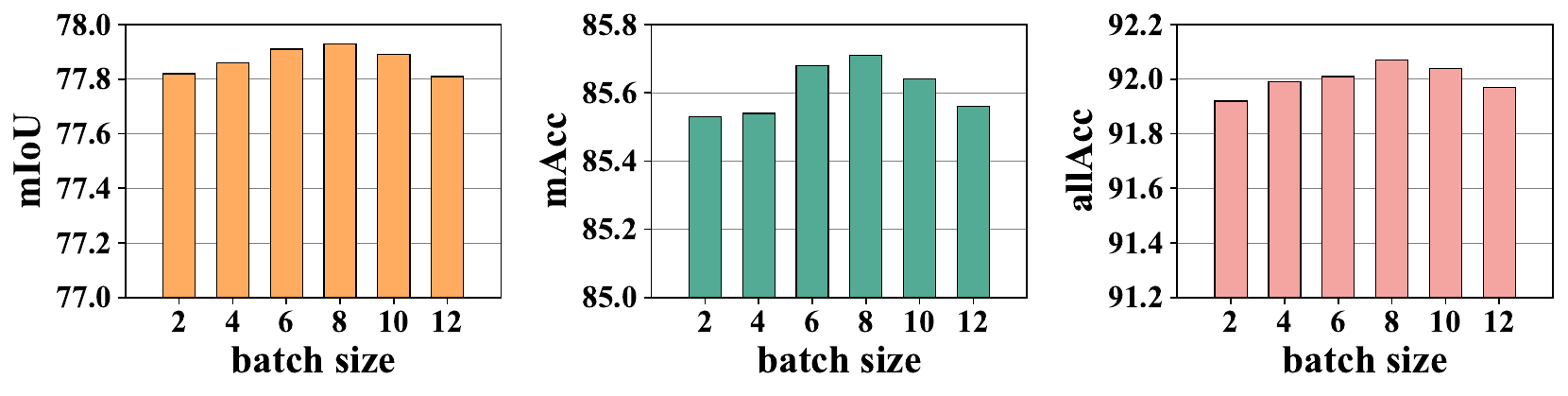}
 \vspace{-0.4cm} 
	\caption{The ablation study of mini-batch Sensitivity. }
	\label{fig3}
\end{figure*}

\textbf{Mini-batch Sensitivity.} We further conduct an ablation study on the impact of different batch sizes for the proposed Cross-Sample Mini-Batch Geometry Distillation (CSMBGD). The results are presented in Figure 3. Although minor performance fluctuations are observed, CSMBGD demonstrates robustness and stability with respect to batch size variations. Moreover, a slight performance improvement can be observed as the batch size increases. This indicates that CSMBGD effectively facilitates the alignment of geometric structures across multiple samples, enabling the student model to better capture the teacher model's multi-sample geometric knowledge.

\textbf{Other experiments} about the analyses of the geometric similarity matrix dimension, the computational comparison, and visualization in nuScenes dataset are shown in supplementary material. 

\section{Conclusion} 

In this paper, we proposed SRKD, a structure- and relation-aware knowledge distillation framework designed to tackle the efficiency and scalability challenges in 3D point cloud semantic segmentation. By transferring both geometric and semantic knowledge from a large, high-capacity teacher model to a compact student model, SRKD effectively bridges the performance gap between lightweight and heavyweight architectures. The affinity matrix-based relation alignment module allows the student to capture fine-grained contextual dependencies, while the cross-sample mini-batch construction strategy facilitates the learning of generalized geometric patterns across diverse point cloud instances. Combined with semantic distribution alignment via KL divergence and strong supervision from ground-truth labels, our method achieves state-of-the-art performance with significantly fewer parameters. Experimental results demonstrate that SRKD is not only accurate but also efficient, making it highly suitable for real-world deployment in resource-constrained environments.

{\small
    \bibliographystyle{unsrt}
    \bibliography{ref}
}


\appendix

\section{Supplementary Material}
\subsection*{Analysis of the Geometric Similarity Matrix Dimension}

In our proposed SRKD framework, the cross-sample mini-batch geometry distillation (CSMBGD) module computes a geometric similarity matrix $M_{ij} \in \mathbb{R}^{N \times N}$ for each pair of samples $(i, j)$ in the batch. The matrix quantifies the cosine similarity between point-wise normalized features. To ensure computational feasibility, we set $N = 1024$ (points per sample), which balances the trade-off between capturing sufficient structural detail and managing GPU memory overhead. 

We further observe that increasing $N$ beyond 1024 yields diminishing returns in segmentation performance, while significantly increasing training time and memory consumption due to the $O(N^2)$ complexity of matrix construction and KL divergence loss. Thus, we adopt $N=1024$ as a default across all experiments to ensure scalability without compromising performance.

\subsection*{Feature Dimension Sensitivity}

To further explore the influence of feature dimensionality in geometric similarity matrices, we conduct ablation experiments by varying the feature dimension used for cross-sample mini-batch geometric alignment. The results on the ScanNet validation set are presented in Table~\ref{tab:dimension_variation}.

\begin{table}[h]
\centering
\begin{tabular}{|c|c|c|c|}
\hline
\textbf{dim} & \textbf{mIoU} & \textbf{mAcc} & \textbf{allAcc} \\
\hline
32  & 77.78 & 85.47 & 91.94 \\
64  & 77.87 & 85.61 & 92.05 \\
128 & \textbf{77.93} & \textbf{85.71} & \textbf{92.07} \\
256 & 77.81 & 85.48 & 91.96 \\
\hline
\end{tabular}
\caption{Segmentation performance under different feature dimensions in the geometric similarity matrix.}
\label{tab:dimension_variation}
\end{table}

The results indicate that increasing the dimensionality from 32 to 128 leads to consistent improvements across all three evaluation metrics. This confirms that higher-dimensional features can encode more expressive geometric information, thereby facilitating more effective similarity modeling between cross-sample point clouds. However, performance slightly drops when the dimension is increased to 256, which suggests a trade-off: excessive dimensions may introduce redundancy or amplify overfitting risks due to noise in the feature space.

Moreover, the computational cost associated with constructing and aligning pairwise similarity matrices increases quadratically with the number of dimensions. As noted in Section A, higher dimensionality leads to a significant rise in memory usage and training time. Based on these observations, we adopt 128 as the default feature dimension throughout our experiments to ensure a good balance between segmentation accuracy and computational efficiency.

\subsection*{Computational Comparison}

To evaluate the efficiency of SRKD, we compare training/inference time, trainable parameters, and GPU memory usage across several variants. The results are shown in Table~\ref{tab:time_params} and Table~\ref{tab:memory}.

\begin{table}[h]
\centering
\caption{Training and Inference Time on ScanNet without using TTA, and Trainable Parameters}
\label{tab:time_params}
\begin{tabular}{lccc}
\toprule
\textbf{Model} & \textbf{Train Time} & \textbf{Test Time} & \textbf{Trainable Params} \\
\midrule
CDSegNet & 27h & 112S & 101,387,354 \\
1/2 ptv3 & 23h & 50S & 11,612,436 \\
KL+MSE & 23h & 50S & 11,612,436 \\
KL+MSE+MATRIX & 64h (DIM=128) & 50S & 11,612,436 \\
KL+MSE+CW & 31h & 50S & 11,612,436 \\
\bottomrule
\end{tabular}
\end{table}

\textbf{Interpretation:} Table~\ref{tab:time_params} shows that student models maintain low parameter counts and fast inference without using test-time augmentation(TTA) across all configurations. The KL+MSE+MATRIX variant notably increases training time due to the cost of affinity matrix computations. However, it preserves low test-time latency, making it still suitable for deployment.

\begin{table}[h]
\centering
\caption{GPU Memory Consumption (Batch Size = 8)}
\label{tab:memory}
\begin{tabular}{lcc}
\toprule
\textbf{Model} & \textbf{Train Memory} & \textbf{Test Memory} \\
\midrule
CDSegNet & 12G & 3.5G \\
1/2 ptv3 & 12G & 3.5G \\
KL+MSE & 14G & 3.5G \\
KL+MSE+MATRIX & 14G & 3.5G \\
KL+MSE+CW & 16G & 3.5G \\
\bottomrule
\end{tabular}
\end{table}

\textbf{Interpretation:} Table~\ref{tab:memory} indicates that all SRKD variants exhibit efficient inference memory usage (3.5GB). Training memory grows modestly with the addition of relation-aware or channel-wise distillation, but remains within acceptable limits for modern GPUs.

\subsection*{Visualization on nuScenes Dataset}

\begin{figure}[h]
\centering
\includegraphics[width=\linewidth]{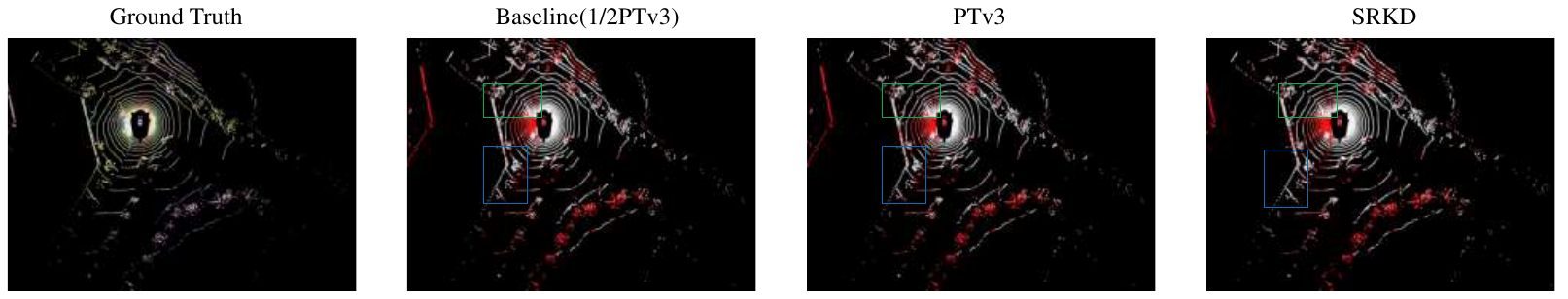}
\caption{Qualitative results on nuScenes. From left to right: ground-truth labels, baseline(1/2PTv3) prediction, 
 PTv3 prediction,SRKD prediction. SRKD improves fine-grained semantic segmentation, especially for small and sparse categories.}
\label{fig:nuscenes}
\end{figure}

To qualitatively assess the segmentation performance on outdoor LiDAR scenes, we provide additional visualizations on the nuScenes dataset in Figure~\ref{fig:nuscenes}. The SRKD-enhanced student model demonstrates more precise boundaries and greater consistency in sparse or occluded regions compared to the non-distilled baseline.

\end{document}